%% file: main.tex
\documentclass{article}
\usepackage{spconf,amsmath,graphicx}
\usepackage{lipsum}
\usepackage{xcolor}
\usepackage{algorithm}
\usepackage{algpseudocode}
\usepackage{booktabs}

\usepackage{enumitem}
\setlist{nosep, leftmargin=14pt}

\usepackage[acronym,nomain,nonumberlist]{glossaries}

\newacronym{map}{mAP}{mean Average Precision}
\newacronym{uq}{UQ}{Uncertainty Quantification}
\newacronym{ece}{ECE}{Expected Calibration Error}
\newacronym{gt}{GT}{Ground Truth}
\newacronym{dece}{D-ECE}{Detection Expected Calibration Error}
\newacronym{iou}{IoU}{Intersection over Union}
\newacronym{pdo}{PDOs}{Patient Derived Organoids}
\newacronym{ls}{LS}{Label-Sampling}
\newacronym{rs}{RS}{Rater-Specific}
\newacronym{lse}{LSE}{Label-Sampling Ensemble}
\newacronym{rse}{RSE}{Rater-Specific Ensemble}
\newacronym{mar}{mAR}{mean Average Recall}

\usepackage{mwe} 

\title{Leveraging Multi-Rater Annotations to Calibrate Object Detectors in Microscopy Imaging}
\name{Francesco Campi$^{1}$, Lucrezia Tondo$^{2,3}$, Ekin Karabati$^{2,3}$, Johannes Betge$^{2,3}$, Marie Piraud$^{1}$}

\address{$^{1}$ Helmholtz AI, Helmholtz Zentrum München, Munich, Germany \\
    $^{2}$Junior Clinical Cooperation Unit Translational Gastrointestinal Oncology and Preclinical Models, \\German Cancer Research Center (DKFZ), Heidelberg, Germany\\
    $^{3}$Department of Medicine II, University Medical Center Mannheim, Medical Faculty Mannheim, \\Mannheim, Germany}
    
\begin{document}
%
\maketitle
\begin{abstract}
\input{1-abstract}

\end{abstract}
\section{Introduction}
\label{sec:intro}
\input{2-Intro}

\section{Methods}
\label{sec:met}

\input{4-Method}

\section{Results}
\label{sec:exp}

\input{5-Results}

\section{Conclusion}
\label{sec:conc}

\input{6-Conclusions}

\section{Compliance with ethical standards}

This study was performed in line with the principles of the Declaration of Helsinki. Approval was granted by the Ethics Committee of the Medical Faculty Mannheim, Heidelberg University (Reference no. 2014-633N-MA and 2016-607N-MA)

\section{acknowledgments}

This work was supported by the Hector Foundation II and Helmholtz AI. 

{\fontsize{10}{11}\selectfont
\bibliographystyle{IEEEbib}
\bibliography{strings,refs}
}


\end{document}

%% file: 1-abstract.tex

Deep learning-based object detectors have achieved impressive performance in microscopy imaging, yet their confidence estimates often lack calibration, limiting their reliability for biomedical applications. In this work, we introduce a new approach to improve model calibration by leveraging multi-rater annotations. We propose to train separate models on the annotations from single experts and aggregate their  predictions to emulate consensus. 
This improves upon label sampling strategies, where models are trained on mixed annotations, and offers a more principled way to capture inter-rater variability. Experiments on a colorectal organoid dataset annotated by two experts demonstrate that our rater-specific ensemble strategy improves calibration performance while maintaining comparable detection accuracy. These findings suggest that explicitly modelling rater disagreement can lead to more trustworthy object detectors in biomedical imaging.

%% file: 2-Intro.tex
In microscopy image analysis deep learning-based object detectors have achieved remarkable detection accuracies, and are widely used to identify and quantify biological structures such as cells, nuclei, or organoids for various applications, including medical diagnostics and drug discovery \cite{multiorg, cell_det, orga_det}. 
Despite their success, they often produce poorly calibrated confidence estimates, and relatively little attention has been paid to how confident the models are in their predictions \cite{Lambert2024}.
This is especially important for microscopy data, which often exhibit substantial variability due to diverse experimental conditions, imaging artefacts, and biological heterogeneity.
Moreover, annotations are provided by human experts, who frequently disagree due to the subjective definitions of target structures and ambiguities from imaging artefacts or noise~\cite{Segebarth2020}.
These inconsistencies in the expert annotations, the so-called \emph{\gls{gt}}, reflect aleatoric uncertainty, a variability inherent to the data itself, which is generally an irreducible component of the overall uncertainty \cite{Hllermeier2020}.
However, deep learning models often struggle to accurately capture it, and explicitly modelling it is essential for developing models that make reliable and trustworthy predictions.

Multi-rater annotations offer an effective way to estimate aleatoric uncertainty by quantifying the level of agreement between different experts \cite{multiorg}.
On one hand, multi-rater annotations help identify the label noise due to annotators mistakes or fatigue, which can also be estimated from the intra-rater agreement of a single rater. 
On the other hand, inter-rater differences capture each rater's individual biases.
In fact, annotations often reflect the subjective interpretation of individual experts. Each rater may focus on different morphological cues or apply slightly different inclusion criteria, leading to systematic but distinct labelling biases.
However, it remains an open question how to effectively leverage multi-rater annotations to train well-calibrated detection models reflecting this inherent uncertainty.

Classification models inherently quantify prediction uncertainty by returning a probability distribution over classes.
A standardized approach to assess the reliability of these confidence scores is \emph{model calibration} \cite{cal_review}. Simply put, a model is calibrated if the predicted confidence matches the likelihood of being correct \cite{Guo2017}. The expected calibration error quantifies this by binning predictions by confidence, measuring the gaps between average accuracy and average confidence in each bin, and aggregating them. 
Deep neural networks tend to be overconfident in their predictions, highlighting the importance of  developing methods able to produce well-calibrated confidence estimates \cite{Guo2017}.
One well-established method for obtaining better-calibrated confidence estimates are Deep Ensembles \cite{deep_ens}. This approach involves training multiple neural networks independently, each  with different random initializations, hyperparameters settings, and data shuffles.
At inference, the models' predictions are aggregated, typically by averaging their probability output \cite{deep_ens}.
Several studies have leveraged multi-rater data for calibration. For example, \cite{dlema} trained deep ensembles for skin-lesion segmentation, where each model learned from different, non-overlapping subsets of the raters' annotations. By contrast, other works randomly sampled annotations from different raters during training~\cite{Jensen2019,Lemay2023}.

In this work, we propose training a  separate model for each rater, allowing each model to learn that rater's individual biases explicitly.
At inference, we then ensemble the rater-specific models by combining their predictions. This strategy not only aggregates complementary perspectives but also provides a natural quantification of the uncertainty arising from annotator disagreement. 
We show that our rater-specific ensembling approach yields better calibration of the model's predictions on organoid bright-field images, without a loss in detection accuracy. 

%% file: 4-Method.tex
The calibration methods discussed above were designed for classification tasks. However, in this work, we focus on object detection, which poses distinct challenges and requires specific adaptations.

\textbf{Calibration of object detectors.}
Unlike classification, where a model outputs a single prediction per sample, object detection models can produce up to hundreds of detections per image. Standard detectors, such as Faster R-CNN \cite{Faster-rcnn}, assign a class confidence score to each detection, making it natural to compute calibration metrics by binning detections rather than data samples, such as whole images. 
Kuppers et. al. \cite{Kuppers2020} evaluated calibration by computing the average precision within each confidence bin, which effectively corresponds to the proportion of detections in that bin that correctly match a \gls{gt} instance.
In addition, they binned detections not only by confidence scores but also by bounding-box position and size, to evaluate model calibration for different image regions and at different scales. However, due to the limited size of our dataset, we restrict our analysis to confidence-based binning to ensure a sufficient number of samples per bin.
We define the \gls{dece} as:
\begin{equation}
    \text{D-ECE} = \sum_{i=1}^M \frac{|B_i|}{N_{det}}\Big|\text{prec}(B_i) - \text{conf}(B_i)\Big|,
    \label{dece}
\end{equation}
where $N_{det}$ is the total number of detections predicted, $B_i$ the $i$-th bin, and $M$ the total number of bins.

\textbf{Deep ensembles of object detectors.}
Ensembling predictions of object detectors requires grouping similar detections \cite{Miller2017}. Consider an image $x$ and a set of $S$ models $\{f^i\}_{i=1}^S$, where each model $f^i$ predicts a different set of boxes for this image $f^i(x) = \{b^i_j\}_{j=1}^{N_i}$. Following Miller et. al. \cite{Miller2017}, we partition all predicted boxes into groups $G_k$, such that for any pair of boxes $(b^i_j, b^{i'}_{j'}) \in G_k$ it holds that (i) they were predicted by two different models $i \neq i'$, and (ii) they have an \gls{iou} above the threshold $\lambda = 0.5$. Next, we compute the confidence score $s$ and coordinates $\mathbf{c}$ of each group by averaging over the detections of that group:
\begin{equation}
    s_{G_k} = \frac{1}{S} \sum_{s^i_j \in G_k} s^i_j, \ \mathbf{c}_{G_k} = \frac{1}{|G_k|} \sum_{\mathbf{c}^i_j \in G_k} \mathbf{c}^i_j
    \label{eq:aggr}
\end{equation}
Note that, unlike \cite{Miller2017}, we average the confidence scores by the ensemble size $S$ (i.e. total number of models) rather than by the group size $|G_k|$. We therefore treat any model that did not predict a box in a given group as if it contributed a confidence score of $0$ to that detection. This ensures that the aggregated confidence reflects the level of agreement across the ensemble, preventing high confidence for detections supported only by a few models.
In contrast, Lyu et al.~\cite{Lyu2021} proposes to keep the confidence score and coordinates of the most confident prediction. However, this aggregation method forfeits the advantages in modelling uncertainty mentioned above.

\section{Experiments}
\label{sec:exp_setup}
Next, we describe the datasets, training pipeline, and evaluation protocol used in our experiments.

\textbf{Datasets and annotation.} 
In this study, we focus on brightfield images of colorectal cancer \gls{pdo}, which are in vitro models that replicate the genetic alterations and spatial organization of the original tumour tissue \cite{org1}. We collected 100 brightfield images of \gls{pdo} cultures at 4$\times$ magnification using a widefield microscope. The goal is to assess organoid viability under different drug treatments by detecting viable \gls{pdo} in the images. This task is particularly challenging because the visual differences between healthy and unhealthy organoids can be very subtle, making them difficult to distinguish.

\begin{figure}[t]

\centerline{\includegraphics[width=\linewidth]{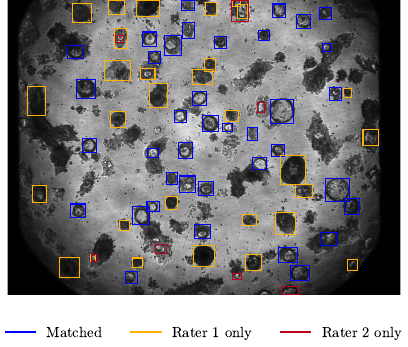}}
\caption{Multi-rater annotations of healthy organoids. Shown are merged detections where raters annotations matched ($\text{IoU} \ge 0.5$), and detections that were annotated by only one rater.}
\label{fig:annotation}
\end{figure}

Two expert raters with comparable levels of experience participated in the annotation process. To facilitate annotation, we first  generated the initial predictions using a pre-trained organoid detector \cite{goat}. The two raters then corrected these predictions using the open-source viewer Napari \cite{napari}. The annotation process was carried out in two phases:
\begin{enumerate}
    \item \emph{Single-rater} dataset: in the first phase, 80 images were randomly sampled and independently annotated by both raters, resulting in two distinct annotations per image.
    \item \emph{Consensus} dataset: in the second phase, the remaining 20 images were annotated jointly by the raters to establish a shared \gls{gt}. 
\end{enumerate}
Figure~\ref{fig:annotation} illustrates the discrepancies between the raters' annotations, highlighting the importance of explicitly modelling aleatoric uncertainty in this setting.

\textbf{Model training and selection.}
Due to the limited amount of annotated data, we fine-tuned the object-detection head of the MaskR-CNN model proposed by \cite{goat} rather than training from scratch. 
The consensus dataset was held out exclusively for final evaluation, while the single-rater dataset was split into 65 training images and 15 validation images. We implemented two training strategies.
\begin{enumerate}
    \item We trained \gls{ls} models by, at each epoch, randomly sampling the annotations from one of the raters for each image.
    \item We trained \gls{rs} models by using only the annotations from either rater 1 or rater 2.
\end{enumerate}
For each training strategy, we explored 80 combinations of hyperparameters, varying the learning rate, batch size, and data augmentation.  
Prior work \cite{deep_ens} has used different random weight initializations to encourage model diversity. In our case, since all models were fine-tuned from the same pre-trained checkpoint, this was not applicable. 
Instead, we varied hyper-parameter settings and data shuffles to increase model diversity. For each training strategy, we selected the best models based on their validation \gls{map}.
Specifically, we took the best 20 \gls{ls} models and the best 10 \gls{rs} models per rater.
Finally, we performed 100 bootstrap resamples of the test set and analysed the resulting \gls{map} distribution to assess whether differences in model performance were statistically significant, as proposed by \cite{rank}.

\textbf{Ensembling.}
We constructed ensembles using two different strategies.  
First, following the approach of \cite{Jensen2019}, we combined the selected \gls{ls} models into a \gls{lse}. Since each \gls{ls} model has been trained on both raters' labels, the ensemble's predictions should represent an intermediate \emph{balanced} view between the two annotators.
We treat \gls{lse} as our baseline.  
Our proposed method is the \gls{rse}, which consists in combining the \gls{rs} models into an ensemble. By aggregating the predictions of models trained on different raters, the ensemble aims to approximate a consensus between them. If each \gls{rs} model has learned its rater's specific biases, then models from different raters will produce contrasting predictions on instances where the annotators disagree. According to Eq.~(\ref{eq:aggr}), such disagreement will cause the ensemble to assign a lower confidence score to that object.
In our experiments, we evaluated different ensemble sizes. For an \gls{lse} of size $S$, we selected the $S$ top-performing \gls{ls} models. For an \gls{rse} of size $S$, we took the top $\frac{S}{2}$ \gls{rs} models for each rater.

%% file: 5-Results.tex
We evaluated the proposed approaches on the colorectal cancer \gls{pdo} dataset described in Section~\ref{sec:exp_setup}.

\begin{figure}[t]
\begin{minipage}[b]{0.49\linewidth}
  \centering
  \centerline{\includegraphics[width=\linewidth]{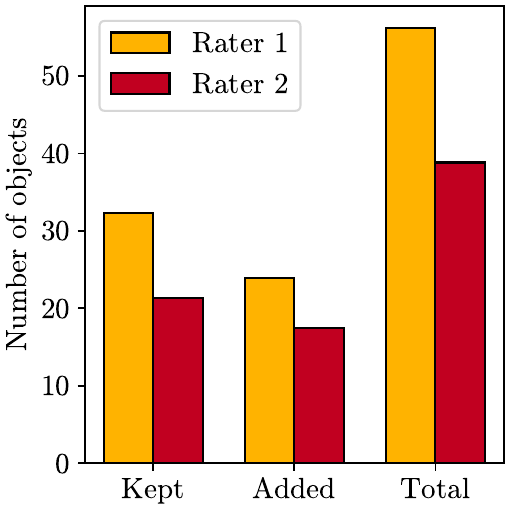}}
  \centerline{(a)}
\end{minipage}
\hfill
\begin{minipage}[b]{.49\linewidth}
  \centering
  \centerline{\includegraphics[width=\linewidth]{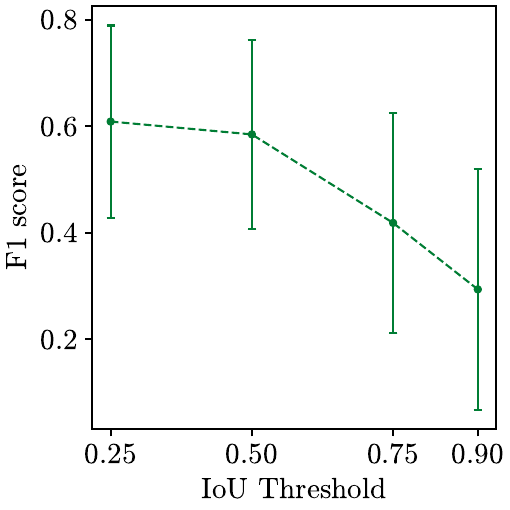}}
  \centerline{(b)}
\end{minipage}

\caption{Inter-rater statistics on the single-rater dataset. (a) Mean number of objects per image retained from the pre-trained model, manually added by each rater, and total number of annotated objects. (b) F1-score between the two raters’ annotations across varying IoU thresholds averaged over all images. Error bars indicate standard deviation.}
\label{fig:inter-rater}
\end{figure}

\textbf{Inter-rater variability and model training.}
First, we compared the annotations produced by the two raters on the single-rater dataset. As shown in Figure~\ref{fig:inter-rater}a, the two raters exhibited distinct annotation styles with Rater~1 annotating about $60$ organoids per image, while Rater~2 only $40$. In addition, Rater~1 retained more objects suggested by the pre-trained model and manually added more compared to Rater~2. This discrepancy is also reflected in the low overall agreement between raters: we observed a mean F1-score of $0.63$ at an \gls{iou} threshold of $0.5$ (Figure~\ref{fig:inter-rater}b). The strong inter-rater differences highlight that aleatoric uncertainty is a significant factor in this biomedical imaging task.
Table~\ref{tab:perf} reports the mean \gls{map} and \gls{mar} over the models trained on the same dataset.
All selected models achieved similar performance on the consensus dataset, and the small standard deviations of these metrics indicate a remarkably low variability in performance across models.
In addition, \gls{map} analysis over 100 bootstrap resampling of the consensus dataset revealed no statistically significant differences between the selected models trained on the same dataset.

\begin{table}[]
    \centering
    \begin{tabular}{ccc}
    \toprule
        Models & \gls{map} $\pm \ \sigma$ & mAR $\pm \ \sigma$ \\
        \midrule
        \gls{ls} &$0.46 \pm 0.0066$ & $0.52 \pm 0.0048$\\
        \gls{rs} (rater 1) & $0.44 \pm 0.0018$& $0.51 \pm 0.0024$\\
        \gls{rs} (rater 2) & $0.45 \pm 0.0015$& $0.51 \pm 0.0015$\\
    \bottomrule
    \end{tabular}
    \caption{\gls{map} and \gls{mar} on the consensus dataset. We report the mean and standard deviation ($\sigma$) across the selected models.}
    \label{tab:perf}
\end{table}

\begin{figure}[t]
\begin{minipage}[b]{.49\linewidth}
  \centering
  \centerline{\includegraphics[width=\linewidth]{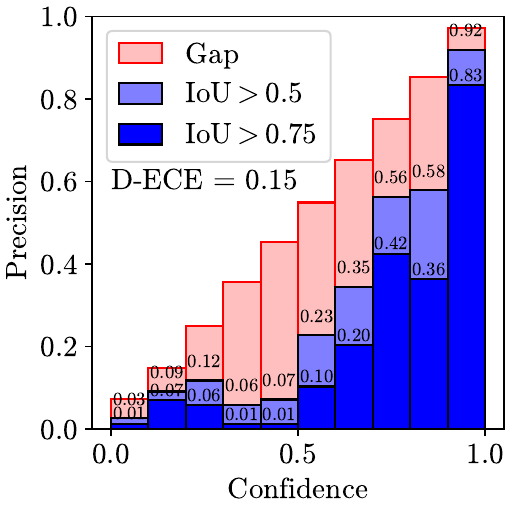}}
  \centerline{(a) \gls{lse}}
\end{minipage}
\hfill
\begin{minipage}[b]{0.49\linewidth}
  \centering
  \centerline{\includegraphics[width=\linewidth]{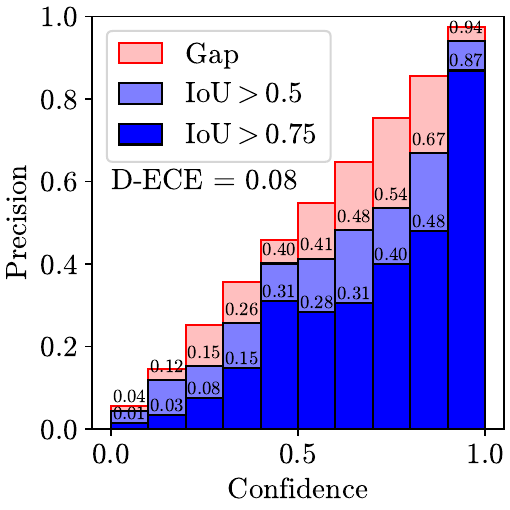}}
  \centerline{(b) \gls{rse}}
\end{minipage}
\caption{Calibration plots for the two ensembling strategies with $20$ models. Blue bars show the average precision observed in each bin $B_i$ computed at \gls{iou} thresholds of $0.5$ and $0.75$, while red bars indicate the gap to a perfectly calibrated model. The \gls{dece} is reported for \gls{iou} $=0.5$. 
}
\label{fig:cal_plots}
\end{figure}

\begin{figure}[t]
\begin{minipage}[b]{.49\linewidth}
  \centering
  \centerline{\includegraphics[width=\linewidth]{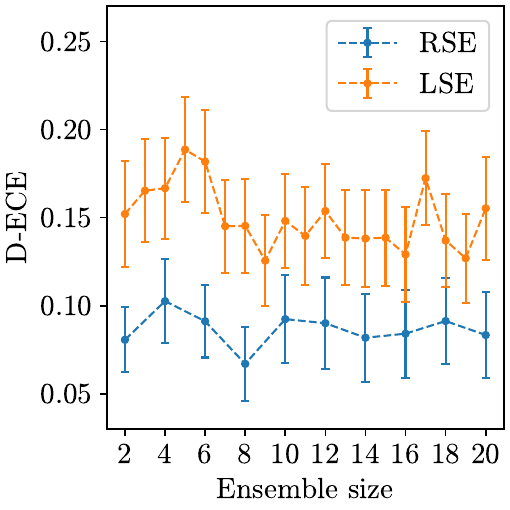}}
  \centerline{(a) \gls{dece}}
\end{minipage}
\hfill
\begin{minipage}[b]{0.49\linewidth}
  \centering
  \centerline{\includegraphics[width=\linewidth]{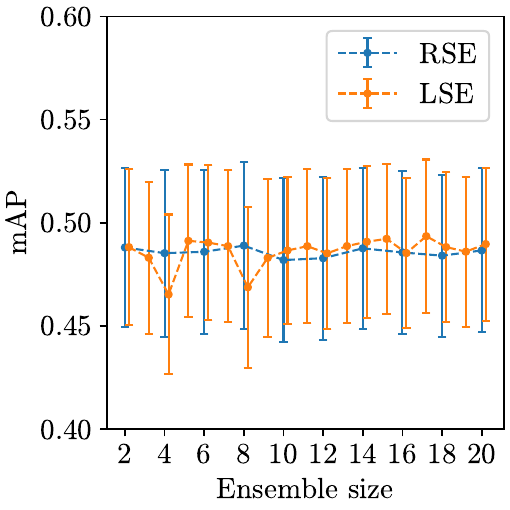}}
  \centerline{(b) \gls{map}}
\end{minipage}
\caption{\gls{dece} and \gls{map} of the ensemble models for different ensemble sizes. Results are averaged over 100 bootstrap iterations; error bars represent standard deviation.}
\label{fig:progress}
\end{figure}

\textbf{Ensembling strategies comparison.}
Next, we investigated how the choice of the ensembling strategy (\gls{lse} versus \gls{rse}) affects model calibration. In Figure~\ref{fig:cal_plots}, we show the calibration plots of the largest ensembles we evaluated for the two strategies ($20$ models). Specifically, at \gls{iou}$=0.5$ the \gls{lse} strategy yielded a \gls{dece} of $0.15$ , while \gls{rse} achieved a \gls{dece} of $0.08$ (a lower \gls{dece} value means a better calibrated model). The calibration plots in Figure~\ref{fig:cal_plots} indicate that \gls{lse} is more overconfident than \gls{rse} for both \gls{iou} thresholds of $0.5$ and $0.75$, exhibiting a larger gap between predicted confidence and precision especially at intermediate confidence levels.
In Figure~\ref{fig:progress}a, we compare how the ensembles performed at different ensemble sizes. We bootstrapped the test set 100 times and visualized the mean and  standard deviation of the \gls{dece} for each ensemble size. 
The \gls{rse} performed significantly better than \gls{lse} for all ensemble sizes.
Finally, Figure~\ref{fig:progress}b, shows the \gls{map} of the ensembles, averaged over the same 100 bootstrapped test sets. The \gls{map} remained essentially constant across ensemble sizes and is similar for both ensembling strategies. 

\section{Discussion}
Our results show that \gls{rse} leads to better calibration than \gls{lse}, while maintaining similar detection performance (see Figure~\ref{fig:progress}a). By capturing each annotator's bias, \gls{rse} naturally reflects inter-rater disagreement. The ensemble then assigns a lower confidence to borderline cases, thereby improving model calibration.

Interestingly, for both strategies the \gls{dece} does not decrease significantly as the ensemble size increases (see Figure~\ref{fig:progress}b). This may be due to the very low variability in performance across the individual models (see Table~\ref{tab:perf}).
One potential reason is our training strategy. As all models were fine-tuned from the same pre-trained checkpoint, instead of using different random initializations, the ensemble models may lack diversity, which could hamper the benefits of larger ensembles~\cite{deep_ens}.


Finally, another important consideration is that, while \gls{rse} improves calibration, its computational cost increases linearly with ensemble size $S$, making inference $S$ times more expensive relative to running a single model.
However, the computational costs of \gls{rse} and \gls{lse} are the same.

%% file: 6-Conclusions.tex

In this study, we demonstrated that training separate detectors for each rater and ensembling them (\gls{rse}) produces better-calibrated predictions than ensembling models trained on mixed labels (\gls{lse}), without sacrificing detection accuracy. The main limitations of this work are the small size of the annotated dataset and the use of only two raters, as producing high-quality multi-rater annotations in biomedical imaging is both costly and time-consuming.

Future work will involve extending the analysis to other applications and using larger datasets, which will enable us to train models from scratch with different initializations. 
Moreover, involving a greater number of raters would produce a finer-grained assessment of inter-rater uncertainty, thus providing a more robust estimate of aleatoric uncertainty. 